\documentclass[aps,pre,reprint,10pt,showpacs,nofootinbib]{revtex4-2}
\usepackage{microtype}
\usepackage{graphicx}
\usepackage{subfigure}
\usepackage{caption}
\usepackage{booktabs} % for professional tables
\usepackage[colorlinks, bookmarks=false, citecolor=blue, linkcolor=red, urlcolor=blue]{hyperref}
\usepackage{xspace}
\usepackage[dvipsnames]{xcolor}
\usepackage{natbib}
\usepackage{bm}% bold math
\usepackage{amsmath}
\usepackage{amsfonts}
\usepackage{amssymb}
\usepackage{mathrsfs} %jolie lettres
\usepackage{mathtools}

\begin{document}

\title{Physics-Informed Graphical Neural Network\\ for Parameter \& State Estimations in Power Systems}
\author{Laurent Pagnier\textsuperscript{1} and Michael Chertkov\textsuperscript{1}}
\affiliation{\textsuperscript{1} Program in Applied Mathematics, University of Arizona, Tucson, USA.}

%\email{laurentpagnier@math.arizona.edu}

%\icmlkeywords{Deep Learning, Graphical Neural Networks, Explainability, Power Flows, Physics Informed Machine Learning}

\begin{abstract} 
Parameter Estimation (PE) and State Estimation (SE) are the most wide-spread tasks in the system engineering. They need  to be done automatically, fast and frequently, as measurements arrive. Deep Learning (DL) holds the promise of tackling the challenge,  however in so far, as PE and SE in power systems is concerned, (a) DL did not win trust of the system operators because of the lack of the physics of electricity based, interpretations and (b) DL remained illusive in the operational regimes were data is scarce. To address this, we present a hybrid scheme which embeds physics modeling of power systems into Graphical Neural Networks (GNN), therefore empowering system operators with a reliable and explainable real-time predictions which can then be used to control the critical infrastructure. To enable progress towards trustworthy DL for PE and SE, we build a physics-informed method, named Power-GNN, which reconstructs physical, thus interpretable, parameters within Effective Power Flow (EPF) models, such as admittances of effective power lines, and  NN parameters, representing implicitly unobserved elements of the system. In our experiments, we test the Power-GNN on different realistic power networks, including these with thousands of loads and hundreds of generators. We show that the Power-GNN outperforms vanilla NN scheme unaware of the EPF physics.
\end{abstract}

\maketitle

\section{Introduction}\label{sec:intro}  

Power System (PS) modeling is the primary tool for the system operator to maintain  awareness of the current state of the system. It is also an enabler for critical decision making, especially with regards to overriding automatic system in the times when the system is stressed. Power engineers rely on physical modeling of the system expressed through a balanced set of the so-called {\bf Power Flow} (PF) equations over the graph of the network, originating from the Ohm's laws governing the physics of the electricity flows and also including physical modeling of devices,  such as generators, transformers and loads, and their control systems \cite{kundur1994power,sauer2017power}. Parameters in the PF equations were estimated and updated systematically.  Uncertainty and delays of these updates were accounted for in the {\bf State Estimation} (SE) and {\bf Parameter Estimation} (PE) \cite{1974Schweppe,1989Wu,2011Gomez-Exposito,2019Zhao},  but generally were not causing concerns in the past -- largely because the PSs were overbuilt, guaranteeing a safe ride through almost any possible state, thus requiring infrequent corrections. However, this (last century) status quo has changed dramatically during the last two decades with introduction of renewable resources pushing PSs to their limits. Emergence of the demand response technology, allowing many small users to become active, has also contributed to increase of uncertainty and fluctuations. It was recognized that to keep the system running one needs to improve observability -- in response, Phasor Measurement Units (PMU) technology was developed \cite{2002Phadke} and installed \cite{PMU} massively in majority of the modern power systems during the last two decades. The PMUs record instantaneously and synchronously physical measurements (of voltages and power flows) at the points of the installation. The technology is powerful,  but also expensive, e.g. due to the massive amount of data which needs to be taken, communicated, processed in real time, stored and accessed. As the result, only major nodes of the power system networks are actually monitored in real time. Given uncertainty in the system parameters, fluctuations in the generation and consumption,  limited observability and massive data flows, the \underline{primary challenge} of the power system operation room remains 
-- to generate projected SE of the PS for the period of interest,  say the next 15 min, fast and reliably. This task also comes hand in hand with the \underline{secondary challenge} -- to generate reliable PE within physical models of the power systems, \underline{actual} -- resolving all nodes of the system (up to hundreds of thousands in the Eastern Interconnect of US), or \underline{reduced} -- effective physical models stated in terms of effective power lines
connecting only major nodes of the PS where PMU measurements are available.

Driven by the tremendous progress in the Data Science Disciplines  power system engineers are now looking very actively into the new Machine Learning (ML) approaches for SE and PE. Such approaches are largely split into three categories -- PS-agnostic, PS-informed and hybrid -- with the previous work on the subject reviewed in Section \ref{sec:related_work}.  

The rest of this manuscript, devoted to further development of the hybrid approach, is organized as follows. Section \ref{sec:formulation} is devoted to the technical introduction to SE, PE, Power Flows and the PS model reductions, as well as to expressing the problems in the ML terms. We then discuss details of the PS data generation and introduce our new Power- Graphical Neural Network (Power-GNN) in Section \ref{sec:experiments}. In this Section of the manuscript, we also present details of our experiments with Power-GNN and a vanilla NN used as a physics-blind benchmark. A special emphasis is given to the reconstruction in the demanding regime of partial observability. Section \ref{sec:conclusions} is reserved for conclusions and discussions of the path forward.

\section{Related works} \label{sec:related_work}

\subsection{Physics-Informed Modeling and Machine Learning}

The term ``Physics Informed Machine Learning" (PIML) in reference to ML was coined in the series of Los Alamos ``Physics Informed Machine Learning" PIML-Workshops which took place in 2016, 2018 and 2020. \cite{PIML-SantaFe}. 
The term came in reference to two complementary aspects -- (a) Physics providing an intuitive and often constructive input into design, analysis and implementation of various ML schemes, and (b) when ML helps to solve a physical problem. This manuscript is concerned with the later aspect of PIML. The approach often translates into embedding equations, describing related physics for the application of interest, into a ML scheme.  These ideas originate from early work,  e.g. of \cite{Lagaris_1998}, on tuning a NN to satisfy the output of an equation (algebraic or differential).  Noticeable applications of the methodology came in the context of dynamical equations (ODEs and PDEs), including the so-called Sparse Identification of Nonlinear Dynamics (SINDy) \cite{2016Brunton}, Physics Informed Neural Network (PINN) \cite{2019Raissi} which was also applied to the PS dynamics in \cite{misyris2020physicsinformed}, and Neural ODE \cite{2018Chen} frameworks.  See also most recent review \cite{willard2020integrating},  and references there in, e.g. emphasizing different objectives in combining elements PIML with the state-of-the-art ML models to leverage their complementary strengths. Physics informed NN modeling in applications to learning SE under limited observability was also discussed in \cite{ostrometzky2020physicsinformed}.

\subsection{Graphical Neural Networks and Applications}

Graph (Convolutional) Neural Network (GNN) is a NN which we build making relations between variables in the (hidden) layers based on the known graph associated with an application, e.g. PS. In this regards, GNN is informed, at least in part, about physical laws and controls associated with the power system operations. One of the first GNN references \cite{kipf2016} focuses on citation networks, and then the approach spread into many applications \cite{zhou2019graph} including geo-location \cite{rahimi2018semisupervised}, infrastructure  modeling \cite{2020Cui}, system of interacting particles \cite{sanchezgonzalez2020learning}  and PSs (mentioned in the next Subsection). 

\subsection{State and Parameter Estimation in Power Systems}

{\bf PS-agnostic techniques} have focused on making SE based on historical and (limited) measurement data utilizing various modern tools of ML, such as Feed-Forward Neural Networks (FFNNs) \cite{Silva1993StateFB,2018Mestav,zamzam2019datadriven,2019Zhang}, autoencoders \cite{2014Barbeiro}, and most recently Convolutional Neural Network (CNN) \cite{2019WentingLi} and Graph Neural Network (GNN) \cite{2019Bolz,2019Donon,owerko2019optimal,liao2021review,2019Kim,20Liao,Chen_2020} (some of these in the context of a related problem of fault detection). It is also worth mentioning most recent NeuroIPS 2020 Learning To Run Power System (L2RPS) competition \cite{L2RPS2020}, won by teams utilizing PS-agnostic approaches. {\bf Physics-Informed approaches} to SE, and related problem of PE in static and dynamic models of PS, actual or reduced, state the problem as a regression based on the PF equations \cite{machowski1997power,huang2009application,chiang2011direct,zhou2011calibration,guo2014adaptive,zhou2015dynamic,chen2016measurement,chavan2017identification,wang2017pmu,2018LearningPowerSystemDynamics,ostrometzky2020physicsinformed},  and also extend in the related areas of probing the proximity to instability and helping in design of the corresponding emergency control actions \cite{ghanavati2016identifying,van1998voltage}, optimization and resource allocation \cite{poolla2017optimal,deka2017acc}. Some of these approaches also dealt with the partial observability \cite{2016Deka,2018Deka,2018Park,2020Deka,2018LearningPowerSystemDynamics},  which is especially well pronounced on the lower voltage (distribution) side of the PS. 
The hybrid approach, attempting to mix the best of the two aforementioned approaches (and inspired by universal methodologies \cite{Lagaris_1998,2016Brunton,2018Raissi} suggesting to blend equations with modern ML) 
was also explored in \cite{misyris2020physicsinformed} to estimate the PS dynamics.  
Ability to predict SE and learn physical models, i.e. to make PE, based primarily on the measurements represents an attractive feature of the framework we advance in this manuscript. In addition to applications mentioned above, validated SE and PE are significant for many other tasks of importance for the system instability, protection and energy management reviewed in many classical books on the subject, e.g. \cite{kundur1994power,sauer2017power}. 

\subsection{Reduced Order Models in Power Systems}

Importance of developing reduced order static and dynamic models, providing computationally light equivalent representation of actual PS, was long time recognized as important, see e.g. review \cite{2012Dukic} and extensive historical references therein. We will highlight in this brief Subsection only a small subset of the waste literature on the subject directly related to the so-called Kron reduction methodology \cite{2010Dobson,2013Dorfler} most relevant to the manuscript.

\section{Problem Formulation: State Estimation and Machine Learning}\label{sec:formulation}

In this Section we set the stage for what follows (experiments with ML schemes) and introduce a few notions and notations from the PS operations, PS analysis and PS state and parameter estimations in Sections \ref{sec:PS-operations},\ref{sec:PF},\ref{sec:SE}, respectively. We discuss our model reduction approach in  Section \ref{sec:Kron}, then resulting in the Physics Informed Machine Learning formulations described in Section \ref{sec:PIML}. 

\subsection{Power System Operations}\label{sec:PS-operations}

Power system should be balanced on the time scale of seconds. This means that electricity generation ought to match the consumption (i.e. the combined value of electric loads and electric/heating losses) at any time. Electricity demand varies over a range of scales -- hourly, daily, weekly or seasonal variations are of a concern in different settings. For example,  in an area with a continental climate, winter demand is generally higher than summer demand. This variations may be linked to local specifics,  e.g. dependence on the electric heating, as in France. In warmer areas, such in the Southern regions of US, the summer demand exceed the winter demand, it is mainly due to air-conditioning. Shorter time scale fluctuations of loads, observed on the scale from seconds to minutes, are mainly due to activities of many small appliances. Note also that renewables, such as wind and solar,  even though they contribute generation, puts an additional pressure on the system injecting uncertainty due to wind and solar which needs to be balanced by other generators or batteries.  Operational redistribution of generation between flexible generators is implemented via energy market mechanisms. Basic optimization behind this task, the so-called Optimal Power Flow (OPF), is resolved frequently, in some energy markets as often as once every 3-5 minutes. OPF is a constrained optimization, where in addition to generation constraints (limits on the generation possible at any generator) and thermal constraints (limiting power or current flowing through each line of the power system) we also add the so-called Power Flow (PF) equations, expressing Ohm's laws, and relating injections/consumptions of power at all the nodes of the system to phases, voltages at the nodes and power flows over the power lines.  PF equations are defined and discussed in more details in the following Subsection.

\subsection{Power Flow Equations}\label{sec:PF} 

Consider a Power System (PS) which operates over a grid-graph, ${\cal G}=({\cal V},{\cal E})$, where ${\cal V}$ and ${\cal E}$ are the set on nodes (generators or loads) and set of lines, respectively. PF equations, governing steady redistribution of power over the system, are stated in terms of the complex-valued (also called complex) powers, $\forall i\in{\cal V}:\ {S}_i\equiv p_i+{\bm i} q_i$, and in terms of the complex-valued (electric) potentials, $\forall i\in{\cal V}:\ {V}_i\equiv v_{i}\exp({\bm i}\theta_i)$, where $v_i$ and  $\theta_i$ denote voltage (magnitude) and phase of the potential at the node $i$, and ${\bm i}^2=-1$. In these notations, the PF equations become, see e.g. \cite{kundur1994power,sauer2017power}, $\forall i\in{\cal V}$ \footnote{We simplify a bit and did not include in the PF equations shunt capacitors, associated with nodes of the systems and representing effects of self-inductance and self-conductance of the PS nodes/buses.
}:
\begin{align}
\!\!\!\!\!\!p_i&=\!\!\!\sum_{j;\{i,j\}\in{\cal E}}\!\!\!v_iv_j\Big[g_{ij}\cos\big(\theta_i-\theta_j\big)+b_{ij}\sin\big(\theta_i-\theta_j\big)\Big],\!\!\!\!\label{eq:load_flow1}\\
\!\!\!\!\!\!q_i&=\!\!\!\sum_{j;\{i,j\}\in{\cal E}}\!\!\!v_iv_j\Big[g_{ij}\sin\big(\theta_i-\theta_j\big)-b_{ij}\cos\big(\theta_i-\theta_j\big)\Big],\!\!\!\!\label{eq:load_flow2}
\end{align}
where $b_{ij}$ and $g_{ij}$ are susceptance and conductance of the power lines, $\{i,j\}\in{\cal E}$;
PF Eqs.~(\ref{eq:load_flow1},\ref{eq:load_flow2}) can be pictured as defining implicitly the PF map, ${\bm \Pi}: {\bm S}\equiv ({S}_i|i\in{\cal V})\mapsto {\bm V}\equiv ({V}_i|i\in{\cal V})$
(The map is implicit, because it requires solving the PF equations, which may have no or may have many solutions.) It can also be stated as an (explicit) inverse map, ${\bm \Pi}^{-1}: {\bm V}\mapsto {\bm S}$. To emphasize parametric dependence of the PF map on the graph, ${\cal G}$, and also on the matrix of the power line admittances defined over the graph, ${\bm Y}\equiv {\bm g}+{\bm i}{\bm b}=(g_{ij}+{\bm i} b_{ij}|\{i,j\}\in{\cal E})$, we use the following notations, ${\bm S}={\bm \Pi}^{-1}_{\bm Y}({\bm V})$.

\subsection{State and Parameter Estimations}\label{sec:SE}

SE -- main task in the center of this manuscript -- consists in, given available observations, to predict other relevant physical characteristics of the system.  For example, assuming that the PS network, i.e. ${\cal G}$ and ${\bm Y}$, are known, and that voltages and phases are measured by PMUs at all nodes of the system,  the SE task may be to estimate injected/consumed active and reactive powers.  This task can obviously be resolved by explicit (that is simple) application of the inverse PF map,  ${\bm \Pi}^{-1}$, set by the PF equations. 

In the case of a limited observability, i.e. when some nodes are unobserved the task of SE can be viewed in two complementary ways. We can pose the question of complementing missing power injections/consumptions only at the nodes where voltages and phases are actually measured.  Alternatively,  we can also pose a more challenging version of the SE under limited observability --- in addition to completing measurements at the observed nodes, also to reconstruct injected/consumed powers and also voltages and phases at all nodes of the system. In this manuscript we will focus on the former, less complicated,  but still challenging enough version of the SE.

The problem of PE consists in reconstructing from PMU measurements structure of the PS graph and also line characteristics, i.e. ${\bm Y}$. This question is relevant because parameters of the power line are known to change due to change in the operational conditions, added or removed vegetation and aging.  These changes are generally infrequent but still significant,  e.g. for making accurate state estimations under complete or partial observability. 

\subsection{Reduced Modeling}\label{sec:Kron}

Motivated by the case of partial observability, we are interested to pose the question of finding equivalent models of power systems providing a description similar to the one provided by the PF Eqs.~(\ref{eq:load_flow1},\ref{eq:load_flow2}), however stated over a reduced set of nodes. Given that the PF Eqs.~(\ref{eq:load_flow1},\ref{eq:load_flow2}) are nonlinear the quest of finding an equivalent (reduced) model is nontrivial. One, admittedly phenomenological (empirical) approach may consists in drawing some useful conclusion from the current-voltage version of the problem (and not power-voltage used in the PF setting),  where model reduction can be evaluated explicitly.

Specifically, let us discuss the so-called Kron reduction -- see \cite{2013Dorfler} and references therein. We divide the nodes of the system into two not-overlapping subsets of the observed (labeled $o$) and unobserved (labeled $u$) nodes.  Then injected/consumed currents, ${I}_i={P}_i/{V}_i$ (not powers!) and potentials of the PS are related to each other according to the Ohm's law describing linear relations between currents and potentials (see Supplementary File for details).
Linearity of the relation between the injected/consumed currents and potentials allows to get a closed relation between currents and voltages associated solely with the observed nodes:
\begin{align}\label{eq:Kirch-o}
    {\bm I}^{(o)}={\bm Y}^{(r)}{\bm V}^{(o)}.
\end{align}
It is therefore suggestive to use the graph structure, ${\cal G}^{(r)}\equiv({\cal V}^{(o)},{\cal E}^{(r)})$, associated with the reduced admittance matrix ${\bm Y}^{(r)}\doteq (\{i,j\}| Y_{ij}^{(r)}\neq 0)$, for building an equivalent PF model -- akin Eqs.~(\ref{eq:load_flow1},\ref{eq:load_flow2}), $\forall i\in{\cal V}^{(o)}$, i.e. 
\begin{align}\label{eq:PF-o}
    {\bm S}^{(o)}={\bm \Pi}^{-1}_{{\bm Y}^{(r)}}({\bm V}^{(o)}),
\end{align}
where, ${\bm Y}^{(r)}$ denote the reduced (we can also call it ``equivalent") admittance matrix associated with the effective (not necessarily real) power lines, $\{i,j\}\in{\cal E}^{(r)}$. ${\bm Y}^{(r)}$ is to be guessed, or as we do it in the following, to be learned. 

\subsection{Physics Informed ML}\label{sec:PIML}

Eq.~(\ref{eq:PF-o}) sets up the backbone of the PIML scheme we develop for the case of partial observability. Specifically,  we aim to solve the following parallel SE and PE learning problem, which we call Power-GNN:
\begin{align}\label{eq:PIML}
& \min\limits_{{\bm \varphi},{\bm Y}^{(r)}}  L_{\text{Power-GNN}}\left({\bm \varphi},{\bm Y}^{(r)}\right),\\ \nonumber & L_{\text{Power-GNN}}\left({\bm \varphi},{\bm Y}^{(r)}\right)  \equiv \frac{1}{N |{\cal V}^{(o)}|}
\sum\limits_{n=1}^{N} 
\Big\|{\bm S}^{(o)}_n \\ \label{eq:Power-GNN} & - \Pi_{{\bm Y}^{(r)}}^{-1}\big({\bm V}^{(o)}_n\big)
- \Sigma_{\bm \varphi}\big({\cal V}_n^{(o)},S_n^{(o)}\big)\Big\|^2 + \mathcal{R}({\bm \varphi})\,, 
\end{align}
where the $l_2$-norm, $\|\cdots\|^2$, assumes summation over observed nodes of the PS; $|{\cal V}^{(o)}|$ is the number of the observed nodes; ${\bm S}^{(o)}_n$ and ${\bm V}^{(o)}_n$, with $n=1,\cdots,N$, denote $N$ samples of the vectors of the complex power and of the electric potential, respectively, measured at the observed nodes, $i\in{\cal V}^{(o)}$. Here in Eq.~(\ref{eq:Power-GNN}), $\Sigma_{\bm \varphi}\big({\cal V}_n^{(o)},S_n^{(o)}\big)$ is introduced to represent effects of the part of the system which is not observed; 
${\bm \varphi}$ stands for the vector of parameters representing the hidden part, and $\mathcal{R}({\bm \varphi})=\alpha\|{\bm\varphi}\|^2$ is the  regularization term, chosen $l_2$. 

In the following, and specifically in Section \ref{sec:NN},  we will introduce the Power-GNN scheme implementing optimization (\ref{eq:PIML}). The scheme will be training simultaneously physics-loaded parameters (elements of the admittance matrix, ${\bm Y}^{(r)}$), and physics-blind parameters representing the $\Sigma_{\bm \varphi}\big({\cal V}_n^{(o)},S_n^{(o)}\big)$ function, via a Graphical Neural Networks (GNN) defined over ${\cal G}^{(r)}$, constructed according to the Kron reduction procedure introduced above in Section \ref{sec:Kron}.

\begin{figure*}[t!]
\center
\includegraphics[width=\textwidth]{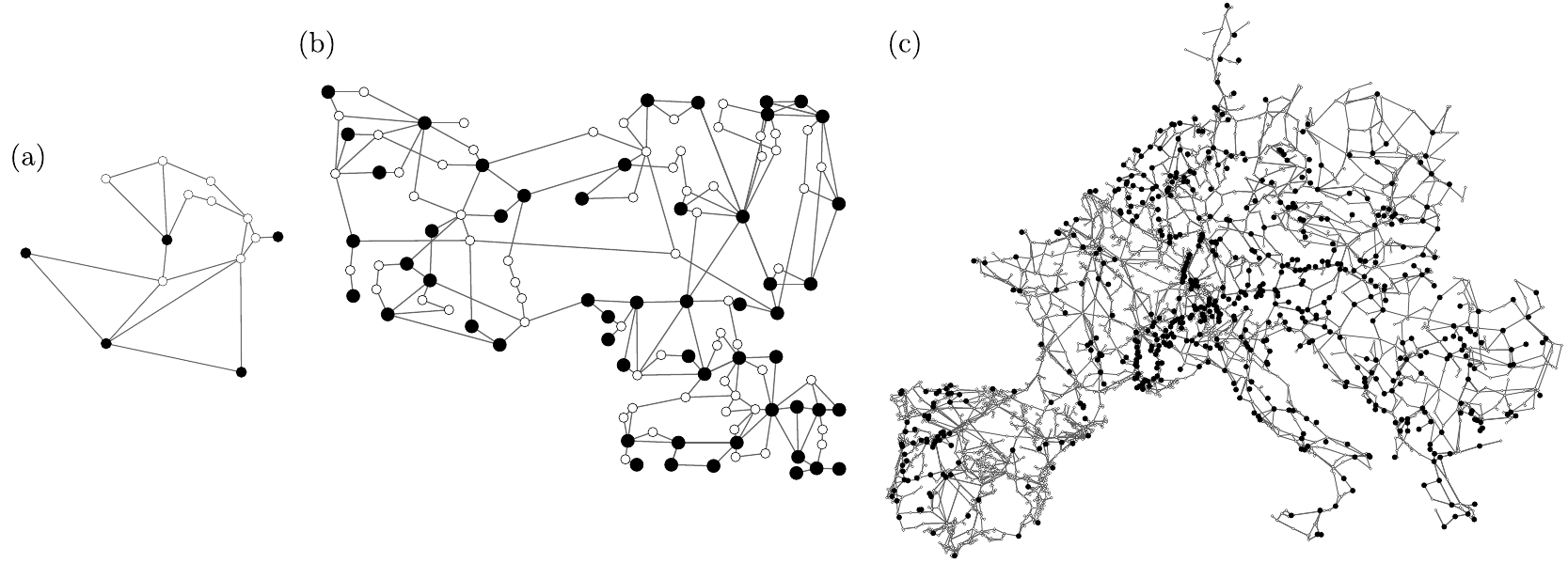}
\caption{(a) IEEE 118-bus test case; (b) IEEE 118-bus test case; (c) PanTaGruEl, a model of the continental European grid \cite{pagnier2019inertia,tyloo2019key}. In each panel, generators and loads are displayed as black and white discs respectively. For partial observability, we assume to have measurements coming only from generators.}\label{fig:models}
\end{figure*}

\section{Experiments}\label{sec:experiments}

We start this Section describing our data generation procedure in Section \ref{sec:data-gen}. We then proceed introducing in Section \ref{sec:NN} our custom physics-informed NN, coined Power-GNN, and a Vanilla NN used as a physics-blind  benchmark. Finally, our experiments, juxtaposing different learning schemes for the set of PS models in different regimes (in terms of observability) are discussed in Section \ref{sec:results}. 

\subsection{Data Generation for testing SE algorithms}\label{sec:data-gen}

To mimic realistic market operations, the PF physics, as well as variability and uncertainty due to fluctuations of loads and renewables, we build the following procedure for generating data samples,  which are used in the following to train the ML models we build for the SE.

We work with three exemplary networks, shown in Fig.~(\ref{fig:models}), representing small, medium and large PSs, and generate synthetic data (to train respective ML models) utilizing a popular OPF solver -- the MATPOWER ~\cite{zimmerman2020matpower}. For each of the networks we generate five data sets that we label as cases \#1-5. (Later  on we will also explain an additional case \#6.)   Our data generation procedure for case \#1 is as follows:
\\  $\bullet$ 
Pick a reference configuration of loads and find respective solution of the OPF problem with the generation cost fixed.
\\  $\bullet$
Devise a family of load configurations each with the same proportion of the loads at different nodes (the same participation factor), but rescaled differently. The rescaling factor is chosen to guarantee  the system reside within the pre-defined interval between the reference (base) load and the peak load. For each load configuration from the family solve the OPF (with the same cost of generation per generator as in the reference case). 
\\  $\bullet$ 
We arrive at the family of samples each represented by complex potentials and injected/consumed powers known for all nodes of the system.\\ 

To produce case \#2 data set we modify procedure injecting random load fluctuation at every node (distributed according to i.i.d. Gaussian distribution with variance equal to a constant fraction of the load at the respective node). For the case \#3 data set active and reactive load are allowed to vary independently. For case \#4, one third of generators is removed from the pool of ``ready to generate'' generators (This relates to the unit commitment problem). A new set of excluded generators is randomly selected for each sample. Finally, for case \#5 data set, the generation cost of generators is randomly chosen in a predefined range, new costs are picked for each sample. Each of the data sets for each of the models consists of 2000 samples.

\begin{table*}[!t]
\caption{Average mismatch of predicted power injections. For the vanilla NN, we also give their values obtained over the training set.}\label{tab:comparison}
\center
\begin{tabular}{l c c c c c c}
\hline
&case \#1 & case \#2 & case \#3 & case \#4 & case \#5 & case \#6 \vphantom{$\sum^i_i$}\\
\hline
Vanilla NN & 4.9E-6 & 7.2E-5 & 6.3E-3 & 5.2E-2 & 6.3E-2 & 1.4E0\\
 & (4.2E-6) & (6.6E-5) & (5.0E-5) & (3.7E-5) & (1.2E-4) & (4.2E-6)\\
Power-GNN & 3.0E-6 & 5.8E-7 & 6.9E-7 & 1.3E-6 & 2.9E-7 & 3.0E-6\\
\hline
\end{tabular}

\end{table*}

We conclude this Subsection with a methodological remark, emphasizing specifics of the datasets representing realistic PS. Actual states of a PS are distributed over respective space in a highly structured way,  which is certainly rather far from uniform. This is due to special correlations induced by the OPF, PF and other procedures, inheriting structure and correlations of the underlying physics and controls. 
This highly nonuniform representation of the phase space makes the PS learning problem more challenging than in other applications, as it requires to go beyond the standard interpolation --- here we ought to extrapolate into regimes which were either unseen or rarely visited.

\subsection{Neural Networks: Methods}\label{sec:NN}

In this brief Subsection we describe two NN schemes -- our custom Power-GNN -- combining elements of the PF equations and GNN,  and -- a standard (one may call it ``vanilla") NN introduced for comparison (as a bench-mark for the Power-GNN). 

\textbf{Power-GNN:} is the algorithm which solves Eq.~(\ref{eq:PIML}) by minimizing the Loss Function (\ref{eq:Power-GNN}) via the gradient descent (we use the Adam algorithm) over the physical parameters, representing the matrix of admittances, ${\bm Y}^{(r)}$ \footnote{We also include in this matrix self-admittances of the nodes associated with shunt-capacitors.}, and also parameters of the Neural Network representing effect of unobserved degrees of freedom on the observed physical characteristics and expressed via $\Sigma_{\bm \varphi}\big({\cal V}_n^{(o)},S_n^{(o)}\big)$ term in Eq.~(\ref{eq:PIML}).
The total number of physical parameters which we learn in the process of training of the PIML model is twice the number of nodes plus twice the number of edges in ${\cal E}^{(r)}$. Even though we call parameters of the NN  -- ${\bm \varphi}$ -- physics blind, we are still attempting to inject in them some physical meaning by building the NN based on the graph, ${\cal G}^{(r)}$, associated with the Kron reduction (see discussion of Section \ref{sec:Kron}). Specifically, we build Graphical Neural Network (GNN) where graph-convolutions, guiding relation between neurons in different layers, follow ${\cal G}^{(r)}$. More info on the construction of the GNN is given below.

\begin{figure}[!h]
\center
\includegraphics[width=0.5\textwidth]{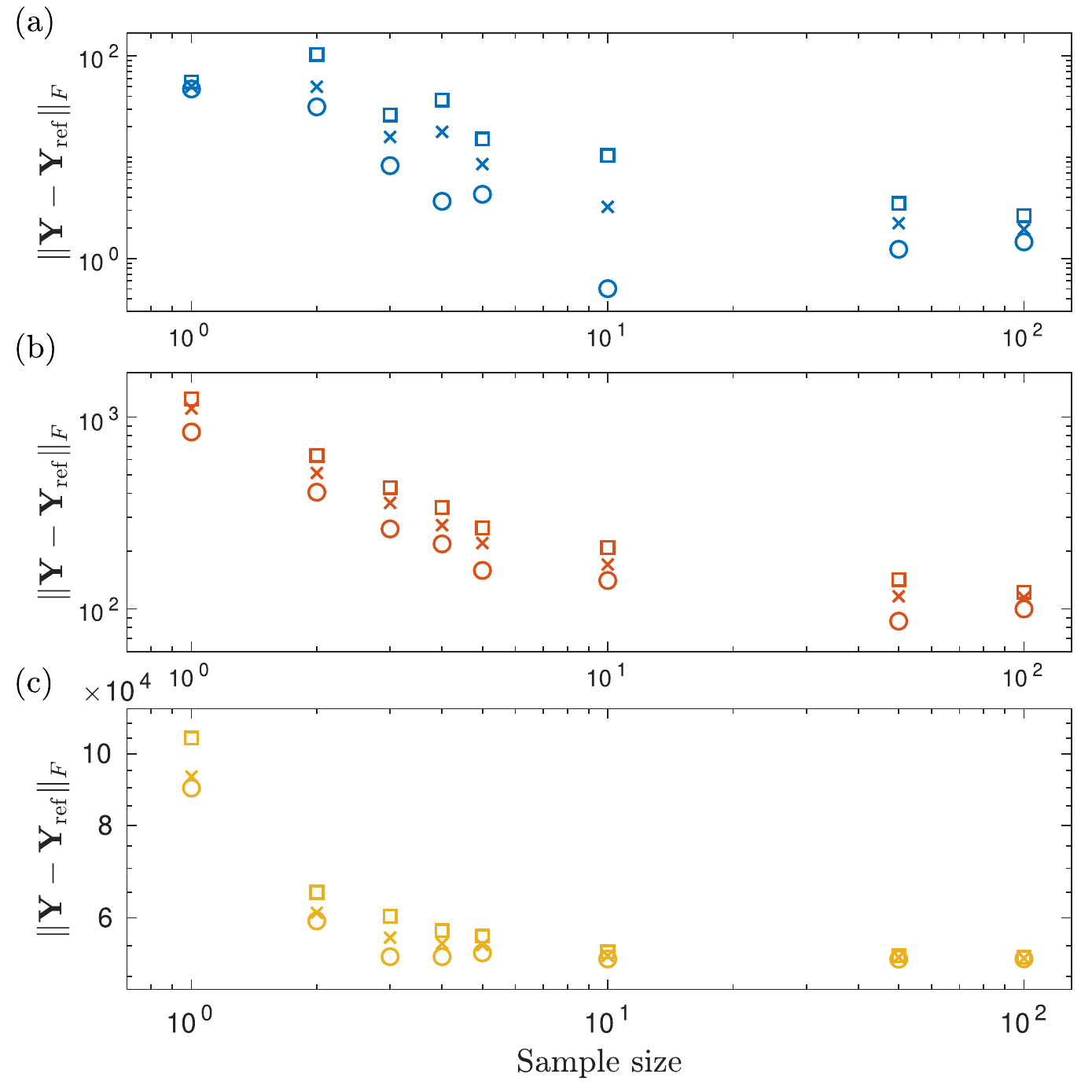}
\caption{Reconstruction of the admittance matrix $\bm Y$ for IEEE 14-bus [panel (a)], IEEE 118-bus [panel (b)] and PanTaGruEl  [panel (c)] models. The min, mean and max values are displayed as circles, crosses and squares respectively (for 10 realizations.) Notice that the quality of the reconstruction delivered by Power-GNN is especially impressive in the case of the large network.}\label{fig:recons}
\end{figure}

\textbf{Vanilla NN:} is presented as a benchmark. We choose a NN with  
four fully-connected (hidden) layers each containing $2\cdot N_{\rm bus}$ neurons. 
We choose this simple architecture after checking, empirically, that it is the smallest layered fully-connected NN which is trained successfully on all the SE data-sets described in Section \ref{sec:data-gen}.  
We tested both RELU and SoftSign activation functions -- and observe similar results. We use the following Loss Function
\begin{gather}\label{eq:LossNN}
    L_{\text{NN}}\doteq \frac{1}{N |{\cal V}^{(0)}|}\sum\limits_{n=1}^{N} \Big\|{\bm S}^{(o)}_n - \text{NN}_{\bm\varphi}({\bm V}^{(o)}_n)\Big\|^2,
\end{gather}
to train the Vanilla NN, mapping samples of the electric potentials to the samples of complex powers at the observed nodes. We call training successful when $L_{\text{NN}}$ reaches $10^{-4}$ on the training set. (Learning rate and number of epochs is chosen to reach the goal. (See supplementary file for  additional details on our choice of the Vanilla NN, including selection of the hyper-parameters.) 

We train both Power-GNN and Vanilla-NN using only fifth of the respective dataset, and then validate it (also to verify that we do not overfit) on the complete dataset.

\subsection{Description and Discussion of the Results}\label{sec:results}

\subsubsection{Full Observability}

\textit{Full observability} means that we assume access to electric potentials and complex powers at every node of the system. This is obviously an ideal (gedanken experiment) setting and later one in the Section we will move on to discuss partial observability where measurements are made over a reduced set of nodes.

Table~\ref{tab:comparison} shows a comparison between Power-GNN and the Vanilla NN benchmark method. Both methods perform similarly when the data sets consists of relatively homogeneous system configurations, as represented in the case \#1. Adding small noise to the values of the injected/consumed complex powers in the case \#2 (in the process of the sample generation, see Section \ref{sec:data-gen} for details) does not alter convergence/performance of either of the two methods compared. However, when we start to increase fluctuations -- case \#3 accuracy of the Vanilla NN starts to decrease.  When we very additionally generator commitment (or altering their cost) -- in cases \#4 and \#5 the accuracy of the Vanilla NN degrades even further. 
We also test on the case \#6 sensitivity of the two schemes to the choice of the reference phase (physics of the PS guarantees that predictions should invariant with respect to shift of all phases of the electric potentials in the system on an arbitrary constant value). Power-GNN passes the test successfully for any values of the phase shift,  while Vanilla NN fails in the case of a sufficiently large shift (of  $20^{\circ}$ or larger).

\begin{figure*}[!t]
\center
\includegraphics[width=\textwidth]{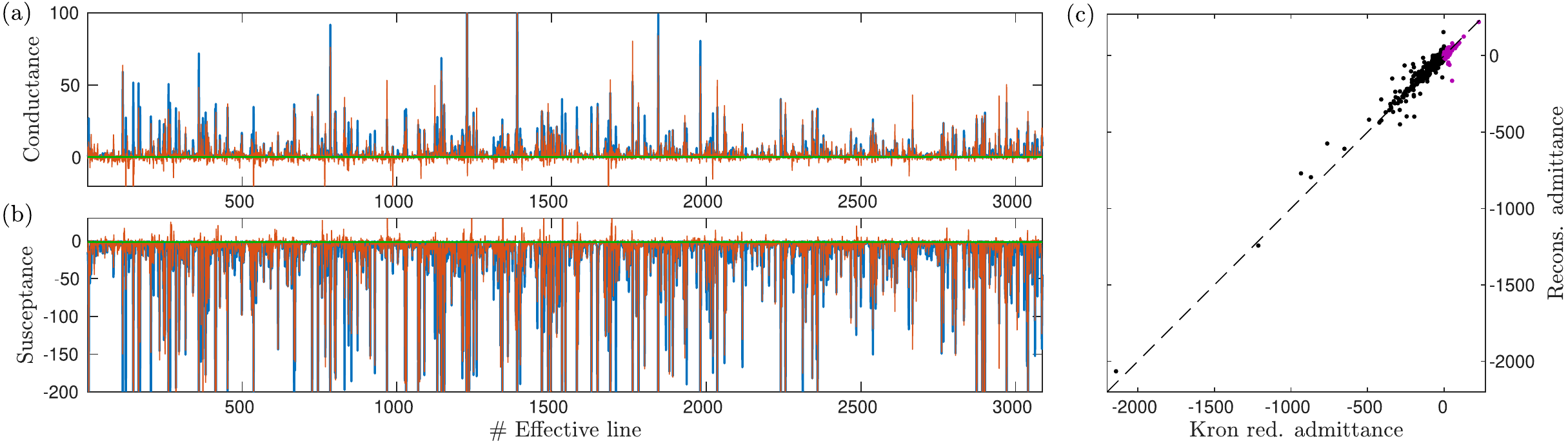}
\caption{Effective line conductances and susceptances, in the reduced model of continental Europe, are displayed in the panels (a) and (b) respectively. Their initial (pre-training) values are displayed in green, while their trained values and their Kron-reduction  counterparts are displayed in red and blue respectively. Panel (c) presents an alternative visualization of the reference-vs-predicted values of the line conductances (purple) and susceptances (black) by showing respective alignment.\label{fig:eff_param}}
\end{figure*}

As mentioned above, the most significant advantage of the Power-GNN is in its ability to  learn the physical, i.e. explainable/interpretable parameters of the PS it is trained on. In Fig.~\ref{fig:recons}, we compare the admittance matrix $\bm Y$ reconstructed through learning with Power-GNN against the actual/reference $\bm Y_{\rm ref}$, following the respective Frobenius/Eucledian norm as the test of the Power-GNN fitness. We observe that a good reconstruction is achieved at an impressively (roughly ten) number of samples. When the number of samples is increased, the reconstruction becomes even more reliable -- as observed through collapse of the difference between the respective min and max values. Normalizing results by the number of nodes show universal asymptotic -- independent on the number of nodes.

\subsubsection{Partial observability}

Transitioning to partial observability, we observe that the Vanilla NN fails dramatically rather fast (with increase in the number of nodes which are not observed). Therefore,  we limit analysis of this regime only to testing performance of the Power-GNN. 

Our main test of the reconstruction quality provided by the Power-GNN consists in comparing impedances of the effective lines predicted by the methods with the respective actual/reference values, which are computed according the Kron reduction, described in Section \ref{sec:Kron}, setting the effective line as valid if amplitude of the respective admittance, $|Y_{ij}^{(r)}|$ exceeds a predefined and system-dependent threshold. 
The results are shown in Fig.~\ref{fig:eff_param}. We observe that the Power-GNN is able to reconstruct the reduced admittance matrix well even in the case of the largest network tested. Notice, that the reconstruction is good, but not perfect. For a small number of lines, we observe values beyond their physically-sensible ranges (positive for conductances and negative for susceptances). We explain this minor deficiency of the Power-GNN by the effect that complexly these miscalculations in the Parameter Estimations (PE) have an insignificant effect on the State Estimations (SE) -- consisting in predicting relations between injected/consumer complex powers and electric potentials at the observed nodes. In other words, these few lines with miscalculated admittances carry much less power than what flows through the rest of the system.

\begin{figure}[!h]
\center
\includegraphics[width=\columnwidth]{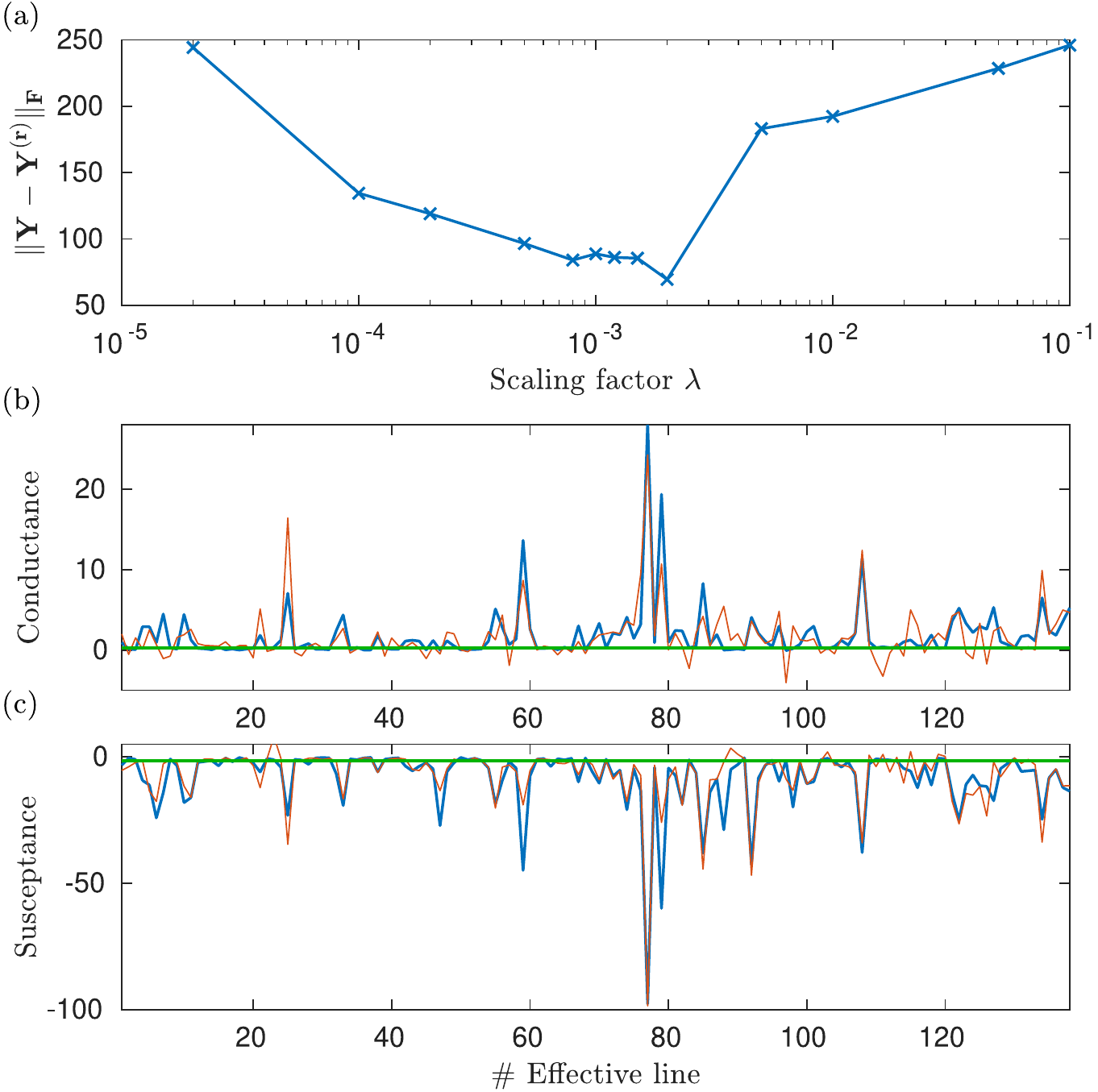}
\caption{Dependence of the quality of the PE reconstruction on regularization [i.e. ${\cal R}(\varphi)$ term in Eq.~(\ref{eq:Power-GNN})], illustrated on the example of the medium-size IEEE 118-bus system. (a) Dependence of the integrated (over all the nodes) quality indicator on the regularization parameter. (b) and (c) shows, respectively, conductances and susceptances of the PS lines reconstructed at the optimal value of the regularization parameter, correspondent to the minimum of the curve shown in panel (a). [(b) and (c) also shows, for reference, respective values the  Kron-reduction counter-parts.] \label{fig:reg}}
\end{figure}

We conclude this Section with a brief discussion of the effect of the NN-regularization, i.e. ${\cal R}(\varphi)=\lambda \|{\bm \varphi}\|^2$ term in Eq.~(\ref{eq:Power-GNN}), on the quality of the SE. We experimented with the scaling factor, $\lambda$, and observed that the SE reconstruction becomes optimal at an intermediate (not small, but also not large) value of $\lambda$,  which we therefore choose empirically, i.e. via experiments, as illustrated in Fig.~\ref{fig:reg}.  

\section{Conclusions and Path Forward}\label{sec:conclusions}

We conclude presenting a brief  summary of the main highlights of the manuscript:
\\  $\bullet$
We address the problem of State Estimation (SE) and Parameter Estimation (PE) in the Power Systems by constructing a novel NN scheme -- Power-GNN.
\\  $\bullet$
Power-GNN borrows physics-informed structure from the Power Flow equations -- main equations power engineers are relying on when monitoring and controlling Power Systems.
\\  $\bullet$
Power-GNN is built to reconstruct SE and PE in the regime of limited observability. To achieve this goal we design a model reduction procedure based on the Kron reduction approach (well known and developed in the PS community). 
\\  $\bullet$
Power-GNN is characterized by both physics-informed parameters (admittances of the effective lines) and physics-blind parameters expressing effect of unobserved degrees of freedom on the observed ones with Graphical Neural Network (GNN).
\\  $\bullet$
Power-GNN is trained by minimizing PF-aware and NN-dependent Loss Function. We utilize PyTorch \cite{PyTorch},  thus taking advantage of analytic differentiation over all the parameters (physics-meaningful and physics-blind).
\\  $\bullet$
Performance of Power-GNN was tested on synthetic power system models of small, medium and large size. We choose Vanilla NN as the physics-blind bench-mark. The performance is validated both in terms of the quality of the LF decay (confirming the quality of the underlying State Estimation), and also in terms of the quality of the Parameter Estimations (reconstruction of the effective line admittances). 
\\  $\bullet$
We show that Power-GNN allows reconstruction of parameters and state estimation in the demanding regime of limited observability, where physics-blind methods are condemned to fail. We argue that Power-GNN returns output in the physical-terms native for the PS practitioners and that it is also capable of extrapolating into regimes of the PS operations which are not contained (or under-represented) in the trained data. 

The following extensions of the methods/approaches and tests presented in the manuscript are in progress:
\\  $\bullet$
Exploring other ways of merging physical and engineering constraints and power system modeling with modern methods in machine learning for better Parameter and State Estimations in PS.
\\  $\bullet$
Testing the methodology on the actual (not synthetic) PMU measurements (subject to collaboration with PS utilities). 
\\  $\bullet$
Generalization of the approach to the Dynamic State and Parameter Estimations, with a special focus on the online learning (returning predictions in real time based on the scale of seconds).
\\  $\bullet$
Extension to active learning,  when SE and PE are also complemented by recommendation to the PS operator on the step mitigating predicted inefficiences and emergencies. 
\\  $\bullet$
Extending the developed methodology to other energy infrastructures, such as natural gas and district heating/cooling system,  and also to related problems (of state and parameter estimation) in other physical infrastructures, e.g. in the domains of the transportation and water-management.

\section*{Broader Impact}

This manuscript contributes our general  approach towards developing methodology for the Physics Informed Machine Learning. As argued above some of the principal steps which were put forward to design the Power-GNN allows extensions to other applications in the domains where observations are partial and straightforward application of otherwise power modern methods of ML does result in  satisfactory predictions.

%\newpage
%\pagebreak
\onecolumngrid
\newpage
\setcounter{figure}{0}
\setcounter{equation}{0}
\setcounter{table}{0}
\setcounter{section}{0}
\renewcommand\thefigure{S\arabic{figure}}  
\renewcommand\theequation{S\arabic{equation}}
\renewcommand\thetable{S\arabic{table}}
\renewcommand\thesection{S\arabic{section}}

\section{More details on the Kron reduction}

As presented in the main text, in the case of partial observability, buses in the system are divided into two complementary sets: one consisting of the observed buses (labelled $o$) and the other of the unobserved buses (labelled $u$). With this reindexing, Ohm's law reads
\begin{equation}
\left[
\begin{array}{c}
\!\!\bm I^{(o)}\!\!\\
\!\!\bm I^{(u)}\!\!
\end{array}\right]
=
\left[
\begin{array}{cc}
\!\!\bm Y^{(oo)}\!\!\ & \!\!\bm Y^{(ou)}\!\!\\
\!\!\bm Y^{(uo)}\!\!\ & \!\!\bm Y^{(uu)}\!\!
\end{array}\right]
\cdot
\left[
\begin{array}{c}
\!\!\bm V^{(o)}\!\!\\
\!\!\bm V^{(u)}\!\!
\end{array}\right]\,,\label{eq:Ohm}
\end{equation}
where $\bm Y^{(oo)}$, $\bm Y^{(ou)}$, $\bm Y^{(uo)}$ and $\bm Y^{(uu)}$ are the submatrices of the admittance $\bm Y$ defined by this reindexing of buses. Solving for $\bm V^{(u)}$ from the second row of Eq.~\eqref{eq:Ohm} and using the obtained expression in the first row, one gets
\begin{equation}
\bm I^{(r)} =\bm Y^{(r)}\,\bm V^{(o)}\,,\label{eq:Kron}
\end{equation}
where $\bm I^{(r)} = \bm I^{(o)} - \bm Y^{(ou)}\bm Y^{(uu)-1}\bm I^{(u)}$ and $\bm Y^{(r)} = \bm Y^{(oo)} - \bm Y^{(ou)}\bm Y^{(uu)-1}\bm Y^{(uo)}$. (For the sake of simplicity and readability, we write $\bm I^{(r)}$ as $\bm I^{(o)}$ in the main text.) In general, the effective current injections $\bm I^{(r)}$ differs from the measured ones $\bm I^{(o)}$. In power system analysis, one usually describes injections in terms of power rather than   with current, hence, from Eq.~\eqref{eq:Kron}, the effective power injections read
\begin{equation}
\bm S^{(r)}\equiv \bm V^{(o)}\circ \bm I^{(r)*} = \bm V^{(o)} \circ \bm I^{(o)*}- \bm V^{(o)}\circ \left(\bm Y^{(ou)}\bm Y^{(uu)-1}\bm I^{(u)}\right)^{*}
=\bm V^{(o)} \circ \left(\bm Y^{\,\rm r}\,\bm V^{(o)}\right)^*\,,\label{eq:Sr}
\end{equation}
where $^*$ denotes the complex conjugate and $\circ$ the Hadamard product (aka the entry-wise product). Rearranging terms in Eq.~\eqref{eq:Sr}, one gets that power injections as they are measured at the observed buses are given by
\begin{equation}
\bm S^{(o)} \equiv \bm V^{(o)} \circ \bm I^{(o)*} = \bm V^{(o)}\circ \left(\bm Y^{(r)}\,\bm V^{(o)}\right)^*+\bm V^{(o)} \circ \left(\bm Y^{(ou)}\bm Y^{(uu)-1}\bm I^{(u)}\right)^*\,.\label{eq:s} 
\end{equation}
The right-hand side of Eq.~\eqref{eq:s} consists of two terms. The first term depends only on the observed voltages and on the reduced admittance matrix $\bm Y^{(r)}$. This term results in Eqs.~(\ref{eq:load_flow1},\ref{eq:load_flow2}) of the main text. The second term is a mixture of observed and unobserved quantities, and therefore it doesn't have a well-defined expression (in terms of observed quantities). This observation motivates us to use a hybrid method consisting of two parts: (a) physics-informed learning of the grid parameters and (b) physics-blind learning of the (second) mixed term. 

\section{Grid Parametrization}

\begin{figure}[!h]
\vspace{-20pt}
\center
\includegraphics[width=0.4\textwidth]{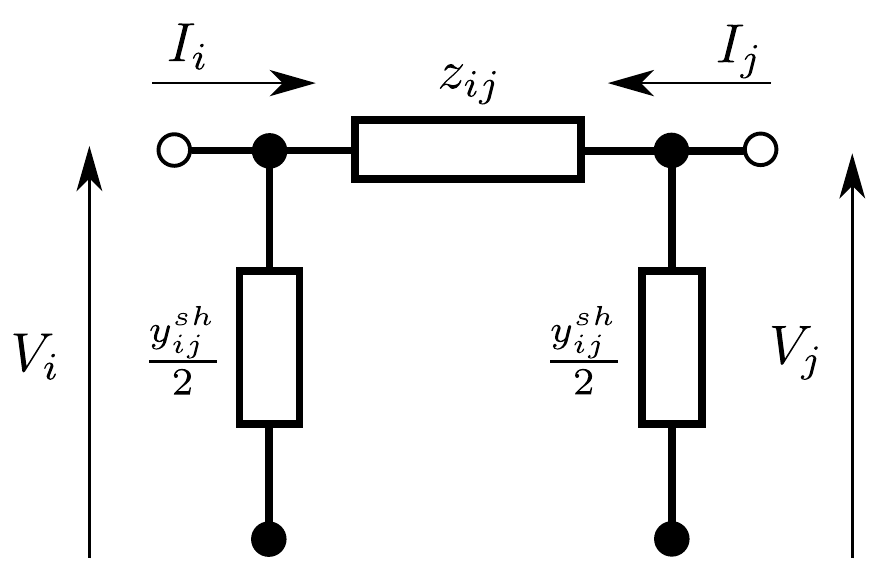}
\caption{Pi-model of a transmission line.}\label{fig:pi_model}
\vspace{-10pt}
\end{figure}

Transmission lines are usually described by an equivalent Pi-circuit model expressed in terms of the line impedance, $z_{ij}=r_{ij}+\bm i x_{ij}$, and two (associated with two end points of the power line) shunt admittances $y_{ij}^{\rm sh}$, see Fig.~\ref{fig:pi_model}. The Pi-circuit model encodes the following relation between currents and potentials at the end-points of the line  
\begin{equation}
\left[
\begin{array}{c}
\!\!I_i\!\!\\
\!\!I_j\!\!
\end{array}\right]
=
\left[
\begin{array}{cc}
\!\!z_{ij}^{-1}+y_{ij}^{\rm sh}/2\!\!\ & \!\! -z_{ij}^{-1}\!\!\\
\!\!-z_{ij}^{-1}\!\!\ & \!\!z_{ij}^{-1}+y_{ij}^{\rm sh}/2\!\!
\end{array}\right]
\cdot
\left[
\begin{array}{c}
\!\!V_i\!\!\\
\!\!V_j\!\!
\end{array}\right]\,.
\end{equation}
The term $z_{ij}^{-1}$ can also be expressed as an admittance $y_{ij}=g_{ij}+\bm i b_{ij}$, where
\begin{align}
g_{ij} &= r_{ij}\big /\big(r_{ij}^2 + x_{ij}^2\big)\,,\\
b_{ij} &= -g_{ij}\big /\big(r_{ij}^2 + x_{ij}^2\big)\,.
\end{align}
One can characterize a line with the combination of its conductance, $g_{ij}$, and susceptance, $b_{ij}$, or equivalently as the combination of its resistance, $r_{ij}$, and reactance, $x_{ij}$. Note however that, in our experiment, we choose to work with the resitance-reactance pairs due to empirical evidence that this choice seems more robust to the (training) initialization. 

\section{Computing infrastructure}

For small and medium size power system models, such as IEEE 14-bus and 118-bus test cases, PowerGNN can be trained on a personal computer within a reasonable time, e.g. training it on IEEE 118-bus test case with $N_{\rm sample}=400$ takes {\footnotesize$\sim$}1600s on a computer with an Intel Core i7 CPU (1.80GHz) and a NVIDIA GeForce MX150 GPU. For larger systems (e.g. PanTaGruEl), training the model within a reasonable time requires the use of an HPC (High-Performance Computer). We have access to the HPC at UArizona with 28 core processors and 46 nodes with Nvidia P100 GPUs. On this HPC, training of the PanTaGruEl model over $N_{\rm sample}=230$ takes {\footnotesize$\sim$}3200s.

\section{List of hyper-parameters}
\vspace{-5pt}
\begin{minipage}{0.45\textwidth}

\captionof{table}{Vanilla NN on IEEE 14-bus test case with full observability\vphantom{$\sum_g$}}
\center
\begin{tabular}{lc}
\hline
Learning rate & 2E-5\\
Number of epochs & 8E5\\
Number of units per layer & 28\\
Number of layers & 3\\
\hline
\end{tabular}

\captionof{table}{Vanilla NN on IEEE 118-bus test case with full observability\vphantom{$\sum_g$}}
\center
\begin{tabular}{lc}
\hline
Learning rate & 2E-5\\
Number of epochs & 8E5\\
Number of units per layer & 238\\
Number of layers & 3\\
\hline
\end{tabular}

\captionof{table}{PowerGNN on IEEE 118-bus test case with full observability\vphantom{$\sum_g$}}
\begin{tabular}{lc}
\hline
Learning rate & 5E-4\\
Number of epochs & 5E4\\
Initial line resistance & 1E-2\\
Initial line reactance & 1E-1\\
Initial bus shunt susceptance & 1E-1\\
Initial bus shunt conductance & 1E-1\\
\hline
\end{tabular}

\captionof{table}{PowerGNN on IEEE 118-bus test case with partial observability\vphantom{$\sum_g$}}
\begin{tabular}{lc}
\hline
Learning rate & 2E-5\\
Number of epochs & 2E4\\
Initial line resistance & 1E-1\\
Initial line reactance & 6E-1\\
Initial bus shunt susceptance & 1E-2\\
Initial bus shunt conductance & 1E-1\\
Number of units per layer & 4E2\\
Number of layers & 3\\
\hline
\end{tabular}

\end{minipage}\hspace{10pt}
\begin{minipage}{0.45\textwidth}
\captionof{table}{Vanilla NN on PanTaGruEl test case with full observability\vphantom{$\sum_g$}}
\center
\begin{tabular}{lc}
\hline
Learning rate & 2E-5\\
Number of epochs & 8E5\\
Number of units per layer & 7618\\
Number of layers & 3\\
\hline
\end{tabular}

\captionof{table}{PowerGNN on IEEE 14-bus test case with full observability\vphantom{$\sum_g$}}
\begin{tabular}{lc}
\hline
Learning rate & 2E-4\\
Number of epochs & 3E4\\
Initial line resistance & 1\\
Initial line reactance & 1\\
Initial bus shunt susceptance & 1\\
Initial bus shunt conductance & 1\\
\hline
\end{tabular}

\captionof{table}{PowerGNN on PanTaGruEl with full observability\vphantom{$\sum_g$}}
\begin{tabular}{lc}
\hline
Learning rate & 2E-5\\
Number of epochs & 5E4\\
Initial line resistance & 1E-2\\
Initial line reactance & 1E-1\\
Initial bus shunt susceptance & 1E-1\\
Initial bus shunt conductance & 1E-1\\
\hline
\end{tabular}

\captionof{table}{PowerGNN on PanTaGrEl with partial observability\vphantom{$\sum_g$}}
\begin{tabular}{lc}
\hline
Learning rate & 2E-5\\
Number of epochs & 5E4\\
Initial line resistance & 1E-1\\
Initial line reactance & 6E-1\\
Initial bus shunt susceptance & 1E-2\\
Initial bus shunt conductance & 1E-1\\
Number of units per layer & 1E3\\
Number of layers & 3\\
\hline
\end{tabular}
\end{minipage}

\end{document}